\documentclass[a4paper, conference]{IEEEtran}
\IEEEoverridecommandlockouts
\usepackage{cite}
\usepackage{amsmath,amssymb,amsfonts}
\usepackage{algorithmic}
\usepackage{graphicx}
\usepackage{textcomp}
\usepackage{xcolor}
\usepackage{url}
\usepackage{makecell}
\usepackage[font=footnotesize]{caption}
\usepackage{cite}
\usepackage{amsmath,amssymb,amsfonts}
\usepackage{algorithmic}
\usepackage{graphicx}
\usepackage[font=footnotesize]{caption}
\usepackage{textcomp}
\usepackage{xcolor}
\usepackage{url}
\usepackage{dot2texi}
\usepackage{tikz}
\usepackage{titlesec}
\titlespacing*{\section}{0pt}{8pt}{4pt}
\titlespacing*{\subsection}{0pt}{6pt}{3pt}
\titlespacing*{\subsubsection}{0pt}{4pt}{2pt}
\usetikzlibrary{arrows, shapes, automata, positioning}
\usepackage{geometry}
\def\BibTeX{{\rm B\kern-.05em{\sc i\kern-.025em b}\kern-.08em
    T\kern-.1667em\lower.7ex\hbox{E}\kern-.125emX}}

\usepackage{hyperref}
\renewcommand{\IEEEkeywordsname}{Keywords}

\begin{document}

\title{A Large-Scale Dataset and Citation Intent Classification in Turkish with LLMs\\
\thanks{This work was supported by TÜBİTAK ULAKBİM.}
}

\author{\IEEEauthorblockN{ Kemal Sami Karaca}
\IEEEauthorblockA{
\textit{Department of Computer Engineering} \\
\textit{ TOBB ETU } \\
Ankara, Türkiye \\
kemalsami.karaca@tubitak.gov.tr}
\and
\IEEEauthorblockN{Bahaeddin Eravcı}
\IEEEauthorblockA{
\textit{Department of Artificial Intelligence Engineering} \\
\textit{ TOBB ETU } \\
Ankara, Türkiye \\
beravci@etu.edu.tr}
}

\maketitle

\begin{abstract}
Understanding the qualitative intent of citations is essential for a comprehensive assessment of academic research, a task that poses unique challenges for agglutinative languages like Turkish. This paper introduces a systematic methodology and a foundational dataset to address this problem. We first present a new, publicly available dataset of Turkish citation intents, created with a purpose-built annotation tool. We then evaluate the performance of standard In-Context Learning (ICL) with Large Language Models (LLMs), demonstrating that its effectiveness is limited by inconsistent results caused by manually designed prompts. To address this core limitation, we introduce a programmable classification pipeline built on the DSPy framework, which automates prompt optimization systematically. For final classification, we employ a stacked generalization ensemble to aggregate outputs from multiple optimized models, ensuring stable and reliable predictions. This ensemble, with an XGBoost meta-model, achieves a state-of-the-art accuracy of 91.3\%. Ultimately, this study provides the Turkish NLP community and the broader academic circles with a foundational dataset and a robust classification framework paving the way for future qualitative citation studies.
\end{abstract}


\section{Introduction}
Citation relationships established among academic publications not only contribute to the advancement of scientific knowledge but also serve as a fundamental metric in evaluating the impact of research outputs. However, traditional metrics overlook the context and purpose of citations, treating all citations as equivalent. In reality, a citation may serve various intentions, such as supporting, discussing, or criticizing a previous work. Identifying the intent behind citations enables more meaningful analyses by capturing these contextual nuances. This study aims to address a significant gap in the Turkish academic literature by developing an artificial intelligence model to classify the intents of citations found in Turkish academic texts.

Understanding citation intent is crucial not only for nuanced research evaluation but also for powering the next generation of intelligent scholarly tools. The automated classification of citation function is a cornerstone for applications that can accelerate scientific discovery, such as semi-automating systematic literature reviews \cite{marshall2019systematicreviews} and building dynamic scientific knowledge graphs that map the conceptual evolution and argumentative structure of a research field\cite{jiang2025evolution_dynamicknowledgegraph}. Such technologies can empower researchers to rapidly identify seminal works, trace the propagation of ideas, and pinpoint contentious research fronts, tasks that are becoming increasingly difficult due to the rapid growth of scientific literature.

Applying these techniques to the Turkish language presents unique and significant challenges. While foundational datasets like SciCite \cite{cohan2019scaffold} have provided a strong basis for this task in English, their models and annotation schemas are not readily transferable. The challenge is twofold. First, Turkish is an agglutinative language characterized by complex morphology and a vast number of word forms derived from a single root, which renders traditional NLP models trained on English corpora ineffective \cite{Oflazer2014TurkishNLPChallenges}. Second, the domain-specific rhetorical conventions embedded in citation practices do not always translate directly across linguistic and academic cultures. These barriers necessitate a ground-up approach, requiring the creation of a new, large-scale, language-specific dataset and the development of models capable of navigating its unique complexities. To execute this, we begin by introducing the first large-scale, publicly available dataset for citation intent classification in Turkish, composed of 2650 annotated examples from Computer Science articles, thereby establishing a foundational resource for the community. Using this dataset, we conduct a comprehensive benchmark of modern Large Language Models (LLMs) like GPT-4o and Gemini, demonstrating the critical performance gains achieved by advancing from manual prompting to an automated prompt optimization framework (DSPy). Finally, to ensure maximum robustness, we develop a high-performance (91.3\% accuracy) stacked ensemble meta-model, establishing a strong and reproducible foundation for future research in Turkish scholarly NLP. The proposed methodology not only enables content-oriented evaluation of Turkish academic publications for the first time but also serves as a potential blueprint for similar analyses in other low-resource languages.
\enlargethispage{\baselineskip}

\section{Related Work}

The field of Citation Intent Classification (CIC) has evolved significantly from early rule-based systems \cite{garzone2000automated, nanba2000classification}, which were often limited in scope. The advent of deep learning led to foundational contributions like the SciCite dataset by Cohan et al. \cite{cohan2019scaffold}, which established a benchmark for training neural models on large-scale data by incorporating contextual features. More recently, the field has shifted towards data-efficient, prompt-based learning with Large Language Models (LLMs). Systems like CitePrompt \cite{lahiri2023citeprompt} have demonstrated that carefully engineered prompts with pre-trained models like SciBERT can achieve high performance with minimal training data, effectively overcoming the need for massive labeled datasets.

In the Turkish academic landscape, CIC research is emergent. Pioneering work by Taşkın et al. analyzed citation patterns in specific fields, proposing custom classification schemes and employing traditional machine learning methods \cite{taskin2018content, taskin2023predatory}. While valuable, these studies were often based on different taxonomies or smaller datasets, highlighting a need for a large, standardized resource to drive progress and enable the use of more powerful, modern techniques.

This methodological evolution is situated within a broader paradigm shift in NLP, driven by the emergent capabilities of LLMs. The introduction of models like GPT-3 demonstrated that large-scale pre-training enables In-Context Learning (ICL), where models can perform novel tasks from just a few examples supplied in a prompt without any parameter updates \cite{Brown_LLM_few_Shotlearner2020}. However, the performance of ICL is critically dependent on the quality of the prompt, leading to a new bottleneck of manual prompt engineering. To address this, recent research has focused on automating prompt discovery. Frameworks like DSPy move beyond simple automation, treating the pipeline as a program to be compiled and optimized, systematically generating and refining prompts to maximize performance on a given task \cite{khattab2024dspy}.

This review of the literature highlights a clear and critical gap: while the global field of CIC has advanced towards sophisticated, prompt-based LLMs, research in Turkish has been constrained by a lack of large-scale, standardized datasets and the continued use of traditional methods ill-suited for the language's morphological complexity. Our work is designed to bridge this divide directly. We address the resource gap by introducing the first large-scale Turkish CIC dataset and tackle the methodological gap by moving beyond manual prompting to implement a state-of-the-art automated optimization pipeline. In doing so, we establish a robust new foundational resource for modern scholarly NLP in Turkish.

\section{Methodology}

\subsection{Dataset Creation and Annotation}
This section details the creation of the first large-scale, publicly available Turkish dataset for citation intent classification in the field of Computer Science. The project was undertaken to address a significant gap in the Turkish academic literature by providing a foundational resource for qualitative citation studies.

\subsubsection{ Source Selection and Citation Extraction }
In this study, the constructed dataset was compiled from academic journals in the field of \textbf{Computer Science} indexed on the \url{https://dergipark.org.tr/} and \url{https://trdizin.gov.tr/} platforms. The aim in selecting these journals was to choose a field where the citation content is explicitly presented and scientific writing standards are relatively homogeneous. The Computer Science discipline offers a suitable sample in this respect, as citations in this field typically appear as methodological, conceptual, or comparative references. In contrast, in disciplines such as social sciences or medicine, citations often appear in footnotes, supplementary paragraphs, or in structurally ambiguous forms, making automatic citation extraction more difficult.

The publications included in the dataset were limited exclusively to research \textbf{articles}. This choice was made to enable comparative analysis with review-type publications that do not contain method sections. It is known that citations in review articles are generally used to provide background information and give less space to citations of experimental methods or results.

For the \textbf{extraction of citation sentences} from designated journals and research articles, the CEX module \cite{opencitations_cex} developed within the GRaSPoS project and utilizing the GROBID library in the background, was used. This module is specifically designed for the precise extraction of citations and contains two models \cite{zenodo_2024_10529646, zenodo_2024_10529709} trained for this task. These models are optimized based on the linguistic structures and citation patterns specific to computer science publications, giving them the ability to distinguish between different citation types with high accuracy. In addition, the CEX module was preferred due to its ability to both generate structured output from the articles in TEI format and provide the data containing citation contexts in JSON format.

\subsubsection{Citation Annotation Scheme }
Datasets used in the task of citation intent classification are typically collected from various academic disciplines and constructed using various labeling schemes. One of the pioneering datasets in this field, \textbf{ACL-ARC}, was gathered from journals in ``Computational Linguistics'' and aimed to categorize citations into six classes to determine their functional aspects \cite{dataset_acl_arc}. Another significant dataset developed for this purpose is \textbf{SciCite}, which was compiled from journals in ``Computer Science \& Medicine.'' It divides citations into three primary classes: BACKGROUND (providing theoretical or contextual information), METHOD (referencing the methods or techniques used), and RESULT COMPARISON (comparing experimental results) \cite{cohan2019scaffold} which was also adopted by the GRaSPoS project.

Furthermore, \textbf{scite.ai}, another tool developed for scientific citation analysis, classifies the content of citations as \verb|supporting|, \verb|contradicting|, or \verb|mentioning|. This classification is concerned not only with the functional aspect of the citation but also with its sentiment, making it possible to evaluate the citation's intent in a positive or negative context \cite{scite_cc}.

This study adopts the five-category citation intent classification system from the Web of Science (WoS) platform: \verb|Background|, \verb|Basis|, \verb|Support|, \verb|Differ|, and \verb|Discuss| \cite{clarivate_cc}. This scheme was chosen for its detailed approach to classifying results: where a comparable system like SciCite uses a single \textit{RESULT COMPARISON} class, the WoS scheme subdivides this concept into \verb|Support| for agreeing findings and \verb|Differ| for conflicting ones. Additionally, the \verb|Basis| class is used to identify the direct application of a referenced methodology. Furthermore, since the WoS database contains sources in both Turkish and English, it offers the potential for future comparative analysis with classifications in both languages.

\subsubsection{The Annotation and Quality Control Process}
To facilitate the labeling process, a web interface was developed for annotations by academic experts \cite{ulakbim_cic}. Participants contributed to the annotation process by logging in either anonymously or with their institutional credentials via this interface. An analysis of the manual labeling results revealed that participants exhibited different approaches, particularly in distinguishing between the \verb|Support| and \verb|Differ| categories. This finding indicates that evaluating citation intent requires a holistic contextual assessment, not just a sentence-level one, and highlights the inherent difficulty in differentiating between semantically close classes like \verb|Support| and \verb|Differ|.

Due to these variations in interpretation among participants, a hybrid approach combining human expertise and AI-assisted automation was adopted for labeling the citation sentences. The foundation of this process consists of manual evaluations conducted by academics from various universities and volunteer users. To support this manual process, complete missing labels, and resolve ambiguous cases that lacked consensus,  In-Context Learning (ICL) with an LLM is utilized. Within this framework, classification predictions generated by Large Language Models (LLMs) such as GPT and Gemini were presented as an assistive reference to the human annotators before making the final decision. This approach aimed to enhance the overall quality and consistency of the dataset.


\subsection{Dataset Statistics}
The dataset was constructed by processing a total of 723 research articles from the Computer Science domain. From these documents, an initial pool of 20,026 citation instances was extracted using the CEX module. Following the annotation and quality control process detailed in the previous sections, a final curated dataset of 2,650 labeled citation sentences was prepared. This final set of 2,650 examples served as the basis for all subsequent modeling and classification experiments presented in this study.

\subsection{Modeling Citation Intent Classification}
Our modeling approach is designed as a multi-stage pipeline that systematically addresses the complexities of citation intent classification in Turkish. The methodology proceeds from a standard ICL method to a more sophisticated, automated prompt optimization framework, and culminates in a robust ensemble model to ensure high-quality and stable predictions.

\subsubsection{In-Context Learning (ICL)}
As a foundational performance benchmark, we employ In-Context Learning (ICL), a technique that leverages the ability of Large Language Models (LLMs) to perform tasks based on a few examples provided directly in the prompt, without any updates to the model's parameters. This approach is ideal for establishing a performance benchmark in a low-resource setting where fine-tuning a dedicated model may be infeasible \cite{kunnath2023prompting}.

Formally, given a citation sentence for classification, $x_{new}$, and a set of $k$ example pairs $(x_i, y_i)$ where $y_i$ is the ground-truth label from the set of five intent categories, we construct a prompt $P = \{(x_1, y_1), ..., (x_k, y_k)\}$. The LLM's task is to predict the label $y_{new}$ by modeling the conditional probability:
\begin{equation} 
y_{new} = \underset{y \in \mathcal{Y}}{\text{argmax}} \ P(y | P, x_{new}) 
\end{equation}
where $\mathcal{Y}$ is the set of all possible intent labels. While powerful, the efficacy of ICL is highly sensitive to the composition of the prompt $P$, including the choice of few-shot examples and the phrasing of the instructional text. This sensitivity motivated our adoption of a more systematic optimization approach.

\subsubsection{Automated Prompt Optimization with DSPy}
To overcome the limitations of manual prompt engineering and discover the most effective prompt structure, we utilize the DSPy framework. DSPy (Declarative Self-improving Python) automates the process of prompt optimization by treating the prompt as a program to be compiled and optimized for a specific task and metric \cite{dspy_khattab2022dsp, dspy_khattab2023dspy, dspy_opsahlong2024optimizing, dspy_soylu2024finetuning,  dspy_singhvi2023dspyassertions}.

\begin{figure*}[htbp]
\centering
\includegraphics[width=0.8\linewidth]{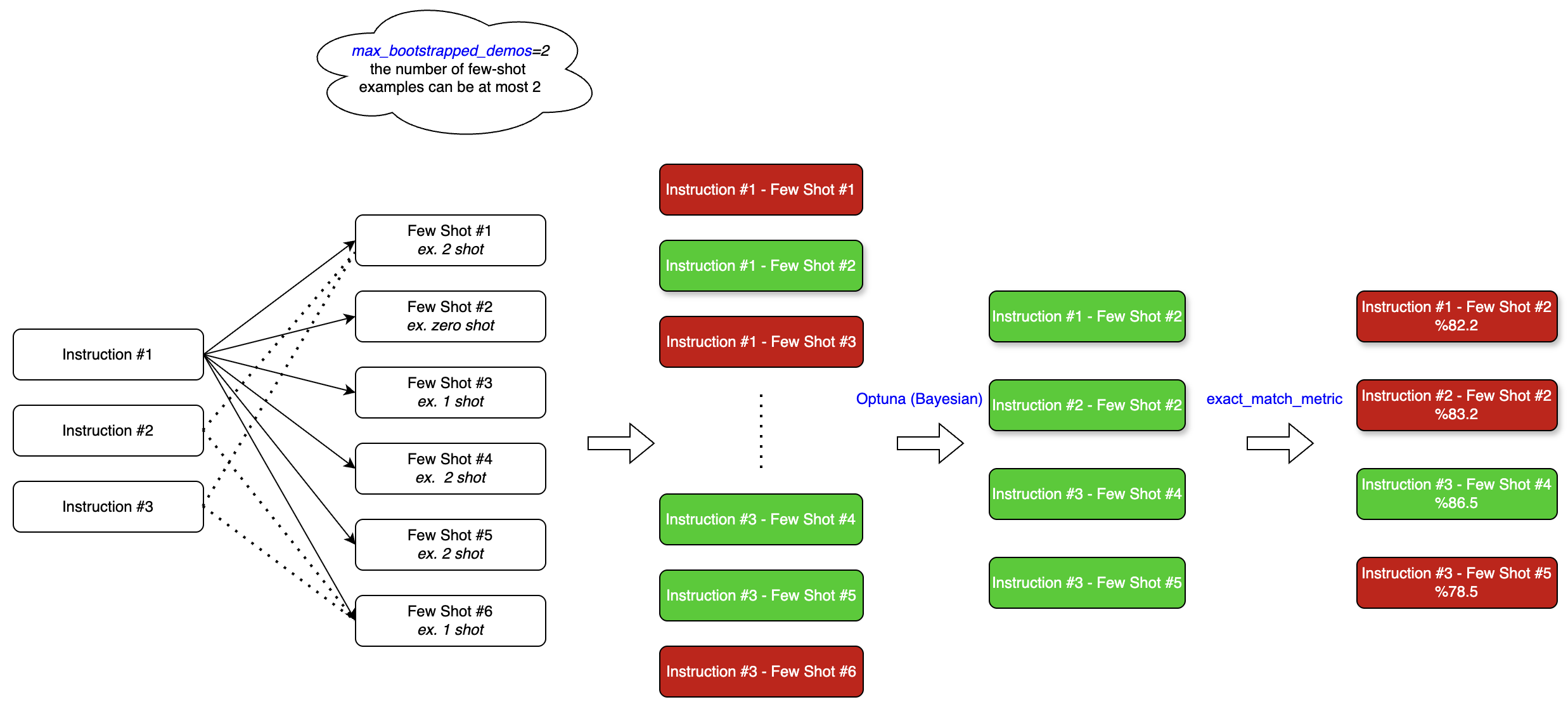}
\caption{DSPy MIPROv2 optimization process}
\label{fig:dspy_mipro}
\end{figure*}

The core objective of DSPy in this study is to find an optimal prompt, $P^*$, that maximizes a scoring function, $S$ (e.g., accuracy), on a validation set, $D_{val}$. The optimization process can be expressed as:
\begin{equation} 
P^* = \underset{P \in \mathcal{P}}{\text{argmax}} \ S(P, D_{val})
\end{equation}
where $\mathcal{P}$ represents the search space of possible prompts. In our implementation, we use the MIPROv2 optimizer within DSPy, which systematically explores variations in both the instructional components and the few-shot demonstration sets. This optimization process is illustrated in \textbf{Fig. ~\ref{fig:dspy_mipro}}. To encourage more explicit reasoning from the LLM during classification, a Chain of Thought (CoT) module is incorporated into the prompt structure, prompting the model to generate intermediate reasoning steps before arriving at a final label.

\subsubsection{Robust Classification via Stacked Ensemble Model}
To enhance classification accuracy and improve the stability of our final predictions, we employ ensemble learning, which combines the outputs from multiple distinct models. The intuition is that by aggregating diverse predictions, the weaknesses of individual models can be mitigated. We explored two primary ensemble strategies.

\textbf{Majority Voting:} A straightforward approach where the final label, $\hat{y}$, is the one predicted by the majority of the $N$ base models. For a given input $x$, the prediction is:
\begin{equation}
\hat{y} = \underset{y \in \mathcal{Y}}{\text{argmax}} \sum_{i=1}^{N} \mathbb{I}(h_i(x) = y)
\end{equation}
where $h_i(x)$ is the prediction of the $i$-th model and $\mathbb{I}$ is the indicator function.

\textbf{Stacked Generalization (Stacking):} A more sophisticated, two-stage approach was adopted for its potential to learn optimal weightings for each base model.

\begin{enumerate}
    \item \textbf{Stage 1 (Base Models):} A set of $N$ diverse base models, $h_1, h_2, ..., h_N$ (representing different LLMs and prompt configurations), are trained or prompted to generate predictions on the dataset.
    
    \item \textbf{Stage 2 (Meta-Model):} The predictions from the base models are used as features to train a meta-model, $H$. For an input sentence $x$, a new feature vector $z = [h_1(x), h_2(x), ..., h_N(x)]$ is constructed. The final prediction, $\hat{y}$, is then produced by the meta-model:
    \begin{equation} \hat{y} = H(z) \end{equation}
\end{enumerate}

For the meta-model $H$, we selected a gradient boosting framework (XGBoost) due to its strong performance in learning from the tabular feature representation created by the concatenated predictions of the base models.

\section {Experiments and Results}

\subsection{Experimental Setup}
For the purpose of reproducibility, this subsection details the specific configurations used in our experiments. We specify the language models employed, the partitioning of our dataset, and the optimization parameters.

\begin{itemize}
    \item \textbf{Models:} The primary Large Language Models (LLMs) utilized for the experiments were OpenAI's GPT-4o, Google's Gemini 2.5 Flash, and Gemini 2.5 Pro.

    \item \textbf{Dataset:} The modeling experiments were conducted on a dataset comprising a total of 2,650 examples. This dataset was partitioned with an 80\% allocation for the training set and 20\% for the validation set.

    \item \textbf{DSPy Configuration:} The automated prompt optimization was performed using the MIPROv2 optimizer within the DSPy framework. The configuration adopted the \texttt{ChainOfThought} (CoT) module to encourage the model to produce intermediate reasoning steps. Key parameters for the optimization included 18 instruction candidates, 9 few-shot candidates, and a total of 27 trials. The detailed parameters for this configuration are summarized in Table~\ref{tab:dspy_config}. 
\end{itemize}

\begin{table}[h!]
\caption{\textsc{DSPy Configuration Parameters}}
\centering
\renewcommand{\arraystretch}{1.2}
\begin{tabular}{|c|c|}
\hline
\textbf{Config Name} & \textbf{Value} \\
\hline
Module & ChainOfThought \\
\hline
Train Set -- Validation Set & \%80 -- \%20 \\
\hline
Instruction candidate & 18 \\
\hline
Few-Shot candidate & 9 \\
\hline
max\_bootstrapped\_demos & 6 \\
\hline
num\_trials & 27 \\
\hline
\end{tabular}
\label{tab:dspy_config}
\end{table}
\subsection{Performance of Individual Models}
This section presents the performance of the individual LLM configurations, starting from the manual baselines and showing the significant improvements gained through automated optimization and shot-selection analysis.

\subsubsection{ICL Performance with Manual Prompts}
The initial performace benchmark was established using In-Context Learning (ICL) with manually crafted prompts. Experiments showed that model performance was highly sensitive to the design of the prompt. An initial prompt version (v000) resulted in an accuracy of 61.3\% for Gemini-2.5-Flash and 52.6\% for GPT-4o. A subsequent revision of the manual prompt (v001) led to a significant performance increase, achieving 83.9\% for Gemini-2.5-Flash and 85.0\% for GPT-4o, demonstrating the critical impact of prompt engineering.

\begin{table}[h!]
\caption{\textsc{Accuracy Values According to Models and Prompts }}
\centering
\small
\renewcommand{\arraystretch}{1.2}
\begin{tabular}{|c|c|c|}
\hline
\textbf{Model} & \textbf{Prompt Version} & \textbf{Accuracy} \\
\hline
Gemini-2.5-Flash & v000 & 0.613 \\
\hline
ChatGPT-4o       & v000 & 0.526 \\
\hline
Gemini-2.5-Flash & v001 & 0.839 \\
\hline
ChatGPT-4o       & v001 & 0.85 \\
\hline
{\textbf{DSPY}} & \multicolumn{2}{c|}{\textbf{0.865}} \\
\hline
\end{tabular}
\label{tab:accuracy_results}
\end{table}

\subsubsection{Performance Improvement via DSPy Optimization}
To address the challenges and performance variance associated with manual prompt design, we employed the DSPy framework to automate the prompt engineering process \cite{khattab2024dspy}. The framework systematically optimized the instructions and the selection of few-shot examples from the training data. This automated approach resulted in a model that achieved a final accuracy of 86.5\%, surpassing the manually tuned prompts and highlighting the effectiveness of the optimization process (Table~\ref{tab:accuracy_results}).

\subsubsection{The Impact of Few-Shot Examples}
An analysis was conducted to evaluate the effect of varying the number of few-shot examples on model performance. The results indicate that more examples do not uniformly lead to better performance. In a zero-shot setting, both Gemini-2.5-Flash and GPT-4o achieved an accuracy of 88.4\%. For Gemini-2.5-Flash, performance peaked at 89.1\% with a single example (1-shot) before declining with additional examples (86.6\% for 2-shot and 87.9\% for 5-shot). Conversely, the performance of GPT-4o degraded as examples were introduced, dropping to 81.4\% (1-shot) and 78.8\% (5-shot).

\subsection{Ensemble Model Performance for Robust Classification}
To obtain more stable and robust outcomes, the classification results from different LLM configurations were aggregated using ensemble strategies. This approach was designed to leverage the strengths of multiple models to create a more accurate and reliable final classifier.

\vspace{5mm} 

\begin{table}[h!]
\caption{\textsc{Accuracy Values According to Models and Shots Counts }}
\centering
\renewcommand{\arraystretch}{1.2}
\begin{tabular}{|c|c|c|c|c|}
\hline
\textbf{Model} & \textbf{Zero-Shot} & \textbf{1-Shot} & \textbf{2-Shot} & \textbf{5-Shot} \\
\hline
Gemini-2.5-Flash & 0.884 & 0.891 & 0.866 & 0.879 \\
\hline
ChatGPT-4o       & 0.884 & 0.814 & 0.816 & 0.788 \\
\hline
Gemini-2.5-Pro   & 0.835 & - & - & - \\
\hline
DSPy             & \multicolumn{4}{c|}{0,865} \\
\hline
\textit{Majority Voting} & \multicolumn{4}{c|}{\textit{ 0.902}} \\
\hline
\textit{Logistic Regression} & \multicolumn{4}{c|}{\textit{0.910}} \\
\hline
\textbf{\textit{Meta Model}} & \multicolumn{4}{c|}{\textbf{\textit{0.913}}} \\
\hline
\end{tabular}
\label{tab:accuracy_shotcounts_results}
\end{table}

A baseline ensemble with majority voting , where the most frequently predicted class was assigned as the final label, achieved an accuracy of 90.2\%. To improve upon this, a stacked generalization (stacking) approach was implemented. This method involves training a meta-model to learn from the outputs of the base classifiers and combine them optimally.

While an initial meta-model using Logistic Regression reached an accuracy of 91.0\%, the best performance was achieved using a gradient boosting-based model (XGBoost). The final stacked ensemble meta-model achieved a headline accuracy of 91.3\%. As shown in Table~\ref{tab:accuracy_shotcounts_results}, this result surpassed the individual models and the simpler ensemble method, demonstrating that integrating model diversity through a learned weighting scheme produces a more robust classifier.

\subsection{Results Analysis and Discussion}
This section interprets the quantitative results, discussing the primary challenges encountered during the study and the implications of the findings.

\subsubsection{The Challenge of Class Imbalance}
\begin{figure}[h!tbp]
\centering
\includegraphics[width=0.7\linewidth]{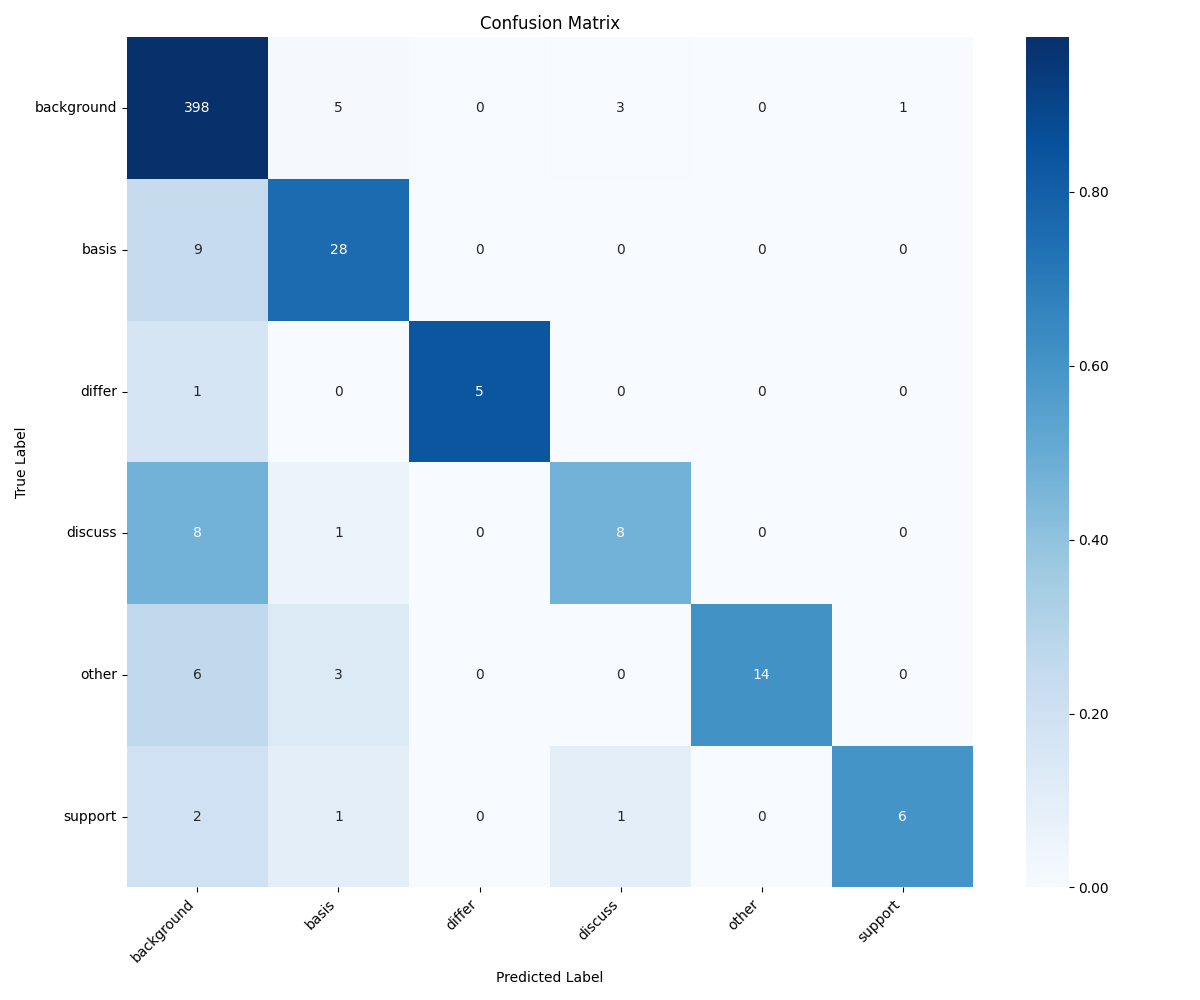}
\caption{Confusion Matrix}
\label{fig:confusion_matrix_normalized_500}
\end{figure}

One of the most significant challenges was the class imbalance within the dataset. As depicted with the confusion matrix in \textbf{Fig. ~\ref{fig:confusion_matrix_normalized_500}}, the analysis revealed that the \texttt{Background} class constituted an overwhelming majority compared to other categories. This distribution is inherent to the nature of academic writing but substantially influenced the model's learning dynamics. The model developed a tendency to overfit the dominant \texttt{Background} class, which negatively impacted its ability to accurately recognize less frequent but semantically important categories like \texttt{Differ} and \texttt{Basis}. Furthermore, this imbalance compromised the few-shot selection process within the DSPy framework, leading to an over-selection of examples from the majority class.

\subsubsection{Qualitative Analysis of Classification Errors}
A key difficulty observed in both manual annotation and automated classification was differentiating between semantically close categories. Human annotators showed varied interpretations, particularly when distinguishing between the \texttt{Support} and \texttt{Differ} classes, highlighting that citation intent evaluation requires a holistic contextual assessment beyond a single sentence. This difficulty was mirrored in the model's performance, which struggled with the accurate identification of underrepresented classes such as \texttt{Differ} and \texttt{Basis}.

\subsubsection{Model Inconsistencies}
During the experiments, it was observed that the LLMs could produce different and inconsistent classifications for the exact same input sentence when fed the same prompt on separate occasions. For instance, a single sentence was classified as \texttt{Basis} in one run and \texttt{Discuss} in another. Similarly, there were cases where an example was labeled as \texttt{support} in one instance and \texttt{basis} in a subsequent run. This observed instability underscored the need for the robust, stacked ensemble approach, which was ultimately used to aggregate predictions and provide more stable and reliable final classifications.

\section{Conclusion}

This study addressed the automatic classification of citation intents within Turkish academic articles, a notable gap in the existing literature. A primary contribution was the creation of the first large-scale, publicly available Turkish dataset for this task, focusing on the Computer Science field. Experiments demonstrated that large language models like GPT-4o and Gemini, paired with In-Context Learning (ICL), can effectively perform classification with few examples. To overcome the performance variability of manual prompting, the DSPy framework was used to automate prompt optimization, successfully increasing model accuracy to 86.5\%. Ultimately, the highest performance was achieved with a stacking-based ensemble model that combined the outputs of different models, reaching a final accuracy of 91.3\% and confirming that integrating diverse models yields a more robust and reliable classifier.

Looking ahead, our analysis points to several promising research directions for advancing this work, primarily aimed at addressing the severe class imbalance uncovered in our dataset. The overwhelming prevalence of the Background class hindered the model's ability to accurately classify semantically critical but less frequent intents like Differ and Basis. One key avenue involves exploring specialized model architectures, such as hierarchical classifiers, to create expert models for more balanced subsets of the data. A second, complementary approach is data-centric, focusing on targeted data collection or leveraging LLMs for synthetic data generation to augment the rare classes. Finally, enhancing the model's input features by incorporating broader contextual information—such as preceding and succeeding sentences or document section metadata—could provide the necessary signals to disambiguate challenging cases. Pursuing these directions promises to yield models with greater robustness and a more nuanced understanding of scholarly discourse.

\bibliography{references}

\end{document}